\documentclass[10pt,twocolumn,letterpaper]{article}

\usepackage{wacv}
\usepackage{times}
\usepackage{epsfig}
\usepackage{graphicx}
\usepackage{amsmath}
\usepackage{amssymb}
\usepackage{subfig}

\wacvfinalcopy 

\def\wacvPaperID{****} 

\ifwacvfinal\fi
\begin{document}

\title{Efficient Action Detection in Untrimmed Videos via Multi-Task Learning}

\author{Yi Zhu and Shawn Newsam\\
University of California, Merced\\
{\tt\small \{yzhu25,snewsam\}@ucmerced.edu}
}

\maketitle
\ifwacvfinal\thispagestyle{empty}\fi

\begin{abstract}
	This paper studies the joint learning of action recognition and temporal localization in long, untrimmed videos. We employ a \textit{ multi-task learning framework} that performs the three highly related steps of action proposal, action recognition, and action localization refinement in parallel instead of the standard sequential pipeline that performs the steps in order. 
	We develop a novel \textit{temporal actionness} regression module that estimates what proportion of a clip contains action. We use it for temporal localization but it could have other applications like video retrieval, surveillance, summarization, etc. 
	We also introduce random shear augmentation during training to simulate viewpoint change.
	We evaluate our framework on three popular video benchmarks. Results demonstrate that our joint model is efficient in terms of storage and computation in that we do not need to compute and cache dense trajectory features, and that it is several times faster than its sequential ConvNets counterpart. Yet, despite being more efficient, it outperforms state-of-the-art methods with respect to accuracy.
\end{abstract}

\section{Introduction}
\label{sec:intro}
The enormous explosion of user-generated videos on the Internet represents a wealth of information. It is essential to understand and analyze these videos for tasks like video recommendation and search, video highlighting, video surveillance, human-robot interaction, human skill evaluation, etc. Automatically localizing and recognizing human actions or events in long, untrimmed videos can save tremendous manual effort and cost.
Much progress has been made on action detection in manually trimmed, short videos over the past several years \cite{yizhu_depth2action_eccvw_2016,15000object2015,Objects2action_Jain_iccv15,actionLocal_context_Soomro_iccv15,actionLocal_track_Weinzaepfel_iccv15,action_tubes_cvpr15_Gkioxari} by employing hand-crafted local features, such as improved dense trajectories (IDT) \cite{idtfWang2013}, or video representations that are learned directly from the data itself using deep convolutional neural networks (ConvNets). However, these approaches are much less effective when applied to long, untrimmed, realistic videos due to the amount of background clutter (non-action clips) as well as the large variation in scene context, motion speed, action style, and camera viewpoint. 

\begin{figure*}[t]
	\centering
	\includegraphics[width=1.0\linewidth,height=5.0in,trim=0 0 0 0,clip]{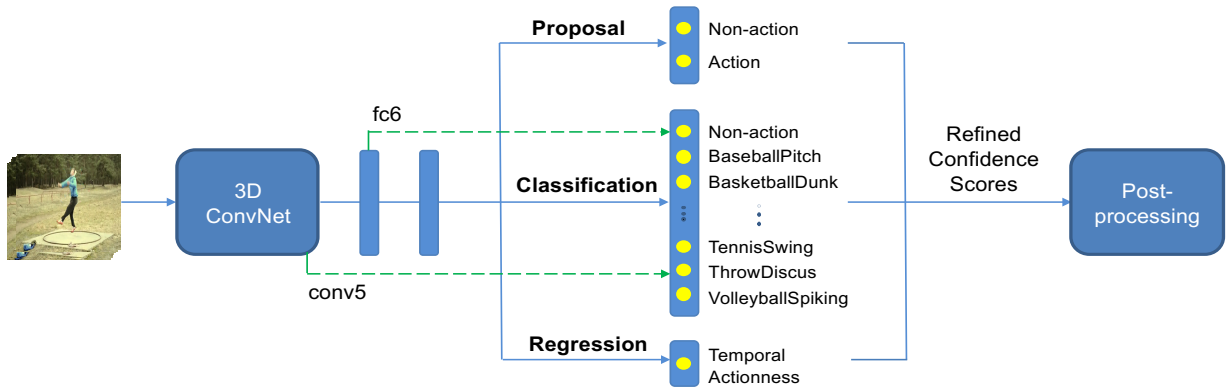}
	\vspace{-46ex}
	\caption{Overview of our multi-task learning framework for action detection. The 3D ConvNet is initialized from \cite{c3d2015}. The three branches of our model simultaneously perform complementary tasks: \textit{Proposal} to determine if a video clip is an action or background, \textit{Classification} of a clip into categories, and \textit{Regression} of what proportion of a clip contains action. The two green dashed lines indicate the multiple intermediate losses used in the action classification branch. The individual outputs are combined to produce refined prediction scores which are used in post-processing to obtain the final detection results.}
	\label{fig:modelConfigure}
	\vspace{-1ex}
\end{figure*}

Presently, the dominant approach to action detection \cite{fast_temporal_action_proposal_fabian_cvpr16,tubelets_cvpr14_jain,yu_FAP_cvpr15,scnn_shou_wang_chang_cvpr16} is a sequential pipeline usually consisting of three steps, either using IDT features plus shallow classifiers or through deep neural networks. First, a video is processed in a temporal sliding window manner to generate action proposals. The candidate proposals are then used to train a more sophisticated action classifier for recognition. Finally, the temporal boundaries of action instances are estimated through post-processing by applying duration priors or non-maximum suppression (NMS). 
While this pipeline has achieved state-of-the-art performance in recent years \cite{learThumos2014,motion_appearance_thumos_wang14,florence_thumos2014,adsc_thumos_15}, we argue that \textit{learning and applying the closely related steps independently has several drawbacks}. First, the performance is upper-bounded by the proposal stage. Second, sequential training prevents the modules from sharing information and restricts the training data each stage sees. Together, these limit the generalizability of the model especially when there is large intra-class variation. Finally, training the stages sequentially is computationally expensive especially for large-scale video analysis.

Inspired by multi-task learning, such as the work on facial landmark detection in \cite{multi_task_face_zhang_eccv14}, we develop a joint learning framework based on 3D ConvNets \cite{c3d2015}. Our goal is to answer the following three questions simultaneously: (1) \textit{Does a video clip contain an action of interest?} (2) \textit{What action class does it belong to?} and (3) \textit{How much temporal actionness does it have?} 
These three problems are closely related and so a joint learning scheme should have better regularization. This will likely improve the network's capacity to discriminate.

Our multi-task learning framework for action detection is outlined in Figure \ref{fig:modelConfigure}. 
The top ``Proposal'' branch is a binary classifier that determines whether a video clip is an action of interest or just background. 
The middle ``Classification'' branch is a conventional action recognition module that classifies a video clip into different action categories. 
This branch combines multiple intermediate losses calculated from last convolutional layer (\textsf{conv5}) as well as the first (\textsf{fc6}) and last (\textsf{fc8}) fully connected layers in order to improve regularization.
The bottom ``Regression'' branch predicts what proportion of a video clip contains the action of interest temporally. This temporal actionness is related to but different from the spatial actionness introduced in \cite{actionness_cvpr14_chen}. That quantity is purely spatial and refers instead to the likelihood that a specific image region contains a generic action instance. 

We show that our joint learning scheme can perform action detection using a single 3D ConvNet. 
In addition, we introduce a data augmentation technique based on shear transformation to increase view-point invariance. 
We also include multiple intermediate losses during training to help prevent overfitting. This strategy is shown to be particularly effective for action recognition in untrimmed videos. Our multi-task network is significantly more efficient, requiring only a fraction of the time of a cascaded ConvNets approach, yet achieves better performance on three popular benchmarks, ranging from THUMOS 2014 (complex) \cite{THUMOS14} to MSR Actions II (simple) \cite{MSR_action_II_PAMI11} and MPII cooking (fine-grained) \cite{MPII_cooking_cvpr12}.

\section{Related Work}
\label{sec:related}
There exists an extensive body of literature on video analysis and human action detection. Here we review only the most related work.

\noindent \textbf{Action Localization:} This field has been studied from two directions: spatio-temporal action localization and temporal action localization. 

Spatio-temporal action localization is an active research area and aims to localize an action simultaneously in space and time. There is a sizable body of work on this problem \cite{DAP3D2016,15000object2015,Objects2action_Jain_iccv15,actionLocal_context_Soomro_iccv15,actionLocal_track_Weinzaepfel_iccv15,action_tubes_cvpr15_Gkioxari,joint_lan_iccv11}; however, most of this work focuses on short, trimmed videos. In addition, spatio-temporal localization requires substantial effort to manually generate the object bounding boxes needed for training. This is not feasible for large-scale video analysis such as unconstrained Internet videos.

In contrast, our focus is on temporal action localization. Given an untrimmed video, we want to determine the temporal bounds of one or more action instances. Many real world applications only need the start and end times of the action/event in a video stream such as in surveillance. Further, temporal action localization can serve as a relatively fast pre-processing step that allows more expensive spatio-temporal localization methods to be applied to just those time periods in which action is detected. 
There has been previous work on temporal action localization.
Yeung et al. \cite{frame_glimpse_yeung_cvpr16} employ a recurrent neural network (RNN)-based agent and use REINFORCE \cite{REINFORCE_1992} to learn the agent's decision policy. Their fully end-to-end approach achieves promising results on two large-scale benchmarks while only needing to observe a fraction of the video frames.
Shou et al. \cite{scnn_shou_wang_chang_cvpr16} adapt 3D ConvNets \cite{c3d2015} for action recognition and design a new multi-stage framework. 
They explicitly take temporal overlap into consideration which benefits the final post-processing step to obtain state-of-the-art results on THUMOS 2014.
Sun et al. \cite{FGA_web_sun_mm15} use noisy tagged web images to discover localized action frames in videos and model temporal information with long short-term memory (LSTM) networks \cite{LSTM_1997}.
Very recently, Heilbron et al. \cite{fast_temporal_action_proposal_fabian_cvpr16} introduced a sparse learning framework to represent and retrieve activity proposals efficiently. 
While our approach shares some structural similarity with these works, it is distinct in that it exploits regularities between closely related tasks for action detection in a multi-task learning framework.

\noindent \textbf{Multi-task Learning:} Multi-task deep networks have become a popular approach to solve multiple semantically related tasks in computer vision. In general, they can be formulated as a fan-out model, where the earlier layers share a single representation and the later layers diverge. 
Multi-task learning has shown promising results in various domains, such as action parsing \cite{DAP3D2016}, semantic segmentation \cite{segment_multitask_cvpr16_dai}, natural language processing \cite{seq2seq_multitask_iclr16_luong}, object detection \cite{object_multitask_iccv15_ghifary}, etc.

Among these works, the DAP3D-Net in \cite{DAP3D2016} is most closely related to our work. They use a general fan-out model to automatically parse actions in videos and  output action types, action attributes, and action locations, simultaneously. However, our work is different from theirs in several key ways: (a) \cite{DAP3D2016} performs spatio-temporal action localization, while our goal is temporal action localization; and (b) they train on well-aligned synthetic action data. In contrast, we utilize widely adopted, untrimmed, realistic video benchmarks.
We take inspiration from \cite{scnn_shou_wang_chang_cvpr16} in designing our multi-task learning framework. But turning a cascaded pipeline into a single network is not easy. In particular, care must be taken to address the data imbalance problem when training a parallel network. 
We also introduce a training data augmentation technique based on shear transformation, design a multiple intermediate loss fusion strategy, and develop a temporal actionness regression module. These contributions highlight the novelty of our joint learning framework compared to previous work on action detection in large-scale untrimmed videos.

\section{Overview}
\label{sec:overview}
\textit{Our goal is to develop a model that can determine the temporal bounds of one or more action instances in a long, untrimmed video.} We particularly want the model to be efficient during training and testing (application) in terms of both computation and storage.

\noindent \textbf{Problem Formulation:} 
For an untrimmed video dataset $\mathbb{V}$, each video $V$ may contain one or multiple action instances $\mathbb{A}$. 
Here, $V$ consists of a set of individual frames $v_{t}$, where $t$ indicates the time index of the frame. 
During training, we generate fixed-length video clips from the annotated videos via temporal sliding windows. These clips are the inputs to our multi-task learning framework. 
The network is trained using five losses: a softmax loss of action/non-action, three softmax losses of action categorization, and a Euclidean squared loss of temporal actionness. 
During prediction, a video is decomposed into clips which are fed to the trained model to obtain per-clip action categorization scores from the last fully connected layer.  These scores are weighted by the output of the other two network branches: the proposal confidence score and the estimated temporal actionness. Finally, non-maximum suppression is applied using the refined action categorization scores, and the top-$k$ results are selected as the temporally localized actions of interest.

\noindent \textbf{Motivation:} 
There are several reasons to develop a multi-task learning framework. First, most existing work does not fully exploit the usefulness of background videos to help train a more robust classifier. Most background videos are discarded after the action proposal stage in order to produce a cleaner and more balanced training data set. Further, as shown in \cite{non_action_wang_cvpr16}, the overall action recognition/detection performance depends on the performance of the action proposal module which is never perfect and will incorrectly label some true actions as background. 
In addition,  the prediction scores of a video clip from the classification module can be high even if it only has a few action frames and the rest are background. This can be problematic because subsequent post-processing steps, such as non-maximum suppression, might remove clips with overall lower action categorization scores but which temporally have a large proportion of action \cite{scnn_shou_wang_chang_cvpr16}. 
We therefore develop a temporal actionness module to quantitatively measure the proportion of action in a video clip. 
In sum, closely related tasks can be integrated in a multi-task learning framework without a reduction in accuracy but with much faster speed during training and evaluation \cite{DAP3D2016,segment_multitask_cvpr16_dai,seq2seq_multitask_iclr16_luong,object_multitask_iccv15_ghifary}. 
Our three main objectives are \textit{efficiency}, \textit{accuracy} and \textit{robustness}.

\section{Methodology}
\label{sec:methodology}
In this section, we first describe our training data augmentation techniques. We then describe our network architecture and the new regression module to predict temporal actionness. Finally, we provide details on post-processing using the refined action categorization scores.

\subsection{Training Data Augmentation}
Most video benchmarks for action detection are extremely small compared to the large-scale image datasets like ImageNet \cite{imagenet_cvpr09}. For example, the largest video dataset with temporal annotations to-date, ActivityNet \cite{activityNet}, only contains 15,410 training action instances. This makes overfitting a challenge in deep learning. Data augmentation techniques, such as random flipping \cite{imagenet_deep_learning_nips12_alex}, corner cropping \cite{wanggoodpractice2015}, multi-scale spatial cropping \cite{sppnet,twostream2014}, and multi-scale temporal sliding windows \cite{scnn_shou_wang_chang_cvpr16}, have therefore been devised to improve the training of deep ConvNets.

In this paper, we utilize three standard data augmentation techniques and introduce a new one for action detection.
We apply multi-scale temporal sliding windows to generate the training video clips. For an untrimmed video, we extract temporal sliding windows of lengths $16$, $32$, $64$, $128$, $256$, and $512$ frames \cite{scnn_shou_wang_chang_cvpr16}. For window lengths $16$ and $32$, the sliding stride is $16$, while for other lengths, it is $32$. This strategy effectively suppresses a bias for shorter windows during NMS \cite{compact_fisher_action_oneata_iccv13}. For each window, we uniformly sample $16$ frames to form our fixed-size input clips. All clips are spatially resized to have a frame size of $128 \times 171$ pixels \cite{c3d2015}.  
During training, we horizontally flip each input video clip with $50\%$ probability. We also do corner-cropping, extracting the four corners and the centers of the frames.

\begin{figure}[t]
	\centering
	\includegraphics[width=1.0\linewidth,trim=0 0 0 0,clip]{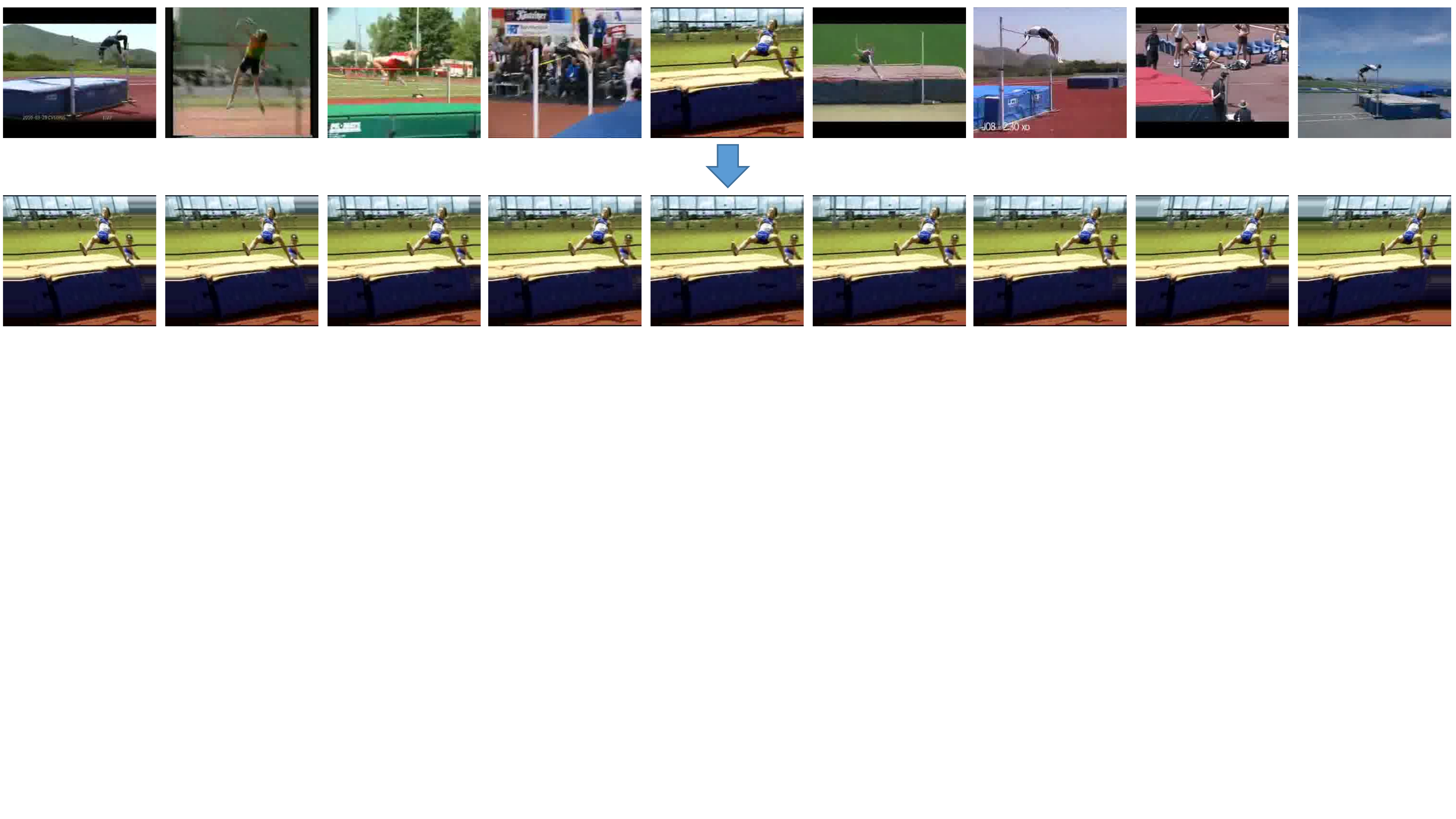}
	\vspace{-20ex}
	\caption{Illustration of our random shear training data augmentation. Top: Sample video frames of action ``High jump'' from the THUMOS challenge 2014 training set.  Bottom: We pick the middle image from the top row and shear it between $-25$ and $25$ degrees for display. While subtle, the difference between the leftmost and rightmost frames does simulate limited viewpoint change and results in better models. This figure is best viewed in color. }
	\label{fig:shear}
	\vspace{-2ex}
\end{figure}

We introduce random shear transformation to simulate viewpoint change for action detection problems. This is motivated by the fact that the same action can appear quite different depending on the viewpoint. For example, sample video frames of the action ``High jump'' are shown in the top row of Figure \ref{fig:shear}. The intra-class variability due to viewpoint change is quite large. The ideal augmentation technique would generate multi-view training clips from a single-view input. This would help reduce overfitting. However, generating 3D videos from 2D is a difficult and computationally expensive problem itself \cite{2dto3d_pose_eccv12_rama,DeepStereo2015,3d_shapenet_cvpr15_wu}, and cannot be done on-the-fly while training deep networks. 

We therefore instead use shear transformations to efficiently simulate limited viewpoint changes. We perform random shearing with a shear parameter $\theta$ between $-25$ and $25$ degrees. This range is determined using cross-validation (too small a value makes no difference while too large introduces unrealistic distortions). Border elements are repeated for padding and the sheared image is cropped to its original size. Sample sheared frames can be seen in the bottom row of Figure \ref{fig:shear}. While subtle, shear transformations are shown to simulate limited viewpoint change. Compare the leftmost and rightmost frames. In the experiments below, we show quantitatively that this augmentation strategy does help prevent overfitting and improves the final detection accuracy on all three benchmarks.

\subsection{Network Architecture}
\label{sec:architecture}
In \cite{c3d2015}, the authors show that 2D ConvNets ``forget'' the temporal information in the input signal after every convolution operation. They propose 3D ConvNets which analyze sets of contiguous video frames organized as clips. 3D ConvNets have shown to be effective at learning spatio-temporal features in video volume data analysis problems \cite{vox2vox_cvprw16_dutran} such as video understanding, optical flow estimation, video coloring, etc.

We therefore base our framework on the 3D ConvNets model C3D described in \cite{c3d2015}. Our network has $8$ convolutional layers, $5$ pooling layers, followed by two fully connected layers, and three parallel outputs. All 3D convolution filters are $3 \times 3 \times 3$ with stride $1$, and all 3D poling layers are $2 \times 2 \times 2$ with stride $1$, except for \textsf{pool1}. In order to preserve temporal information in the earlier layers, \textsf{pool1} has size $1 \times 2 \times 2$. The detailed network structure is shown in Figure \ref{fig:modelConfigure}. 
Since C3D is trained on the large Sports1M video dataset \cite{KarpathyCVPR14}, it serves as a good initialization point for training our model. The inputs to our network are mini-batches of video clips of size $16 \times 3 \times 112 \times 112$, and the outputs are three sets of confidence scores: action or not, action category, and temporal actionness.

In order to reduce overfitting and increase the influence of the action categorization branch, we introduce two additional loss functions which are conceptually similar to the auxiliary classifiers in \cite{inception_v3_2015}. Besides the softmax loss from layer \textsf{fc8}, we add two more softmax losses to regularize the learning, one from layer \textsf{conv5} and another from layer \textsf{fc6}. The benefits of this are two-fold. 
(i) Layer \textsf{conv5} contains weak semantic, but strong spatial information, which is important for motion understanding; layers \textsf{fc6} and \textsf{fc8} can provide high-level semantic and abstract information about the video clip. These three loss functions thus incorporate complementary information from the top layers of the network. 
(ii). Incorporating these additional intermediate losses promotes action categorization as the most important of the three main tasks.
Improved action recognition also benefits the other two branches. 
Training our network thus considers a total of five losses.


Note that, though there is strong motivation to perform temporal action localization via a parallel multi-task framework, the joint learning of the tasks is a challenge due to potential imbalance in the training data. The training data that is optimal for one branch of the network might not be optimal for the others. The action proposal branch is a binary classification problem and seeks to distinguish between two classes, action and non-action. In contrast, the action categorization branch is a multi-class classification problem and seeks to distinguish between $N+1$ classes ($N$ classes plus background). Usually $N > 1$ so if we generate balanced training data for the action proposal branch, it will be unbalanced for the action categorization branch, and vice versa. 
We explore two ways to address this problem. (i) Provide balanced data for the action proposal task and tune the class weights when calculating the action categorization loss. This essentially means giving more attention to action clips when training the categorization branch. Or, (ii) provide balanced data for the action categorization branch and tune the class weights when calculating the action proposal loss.
We show in Section \ref{sec:experiments} that (ii) results in higher accuracy. One possible reason is that action categorization is our main goal, while action proposal is just a supporting task. If we provide balanced data for the action proposal task, our network sees too many non-action examples and learns features more from background video clips than from actions of interest.

\subsection{Temporal Actionness}
Actionness was introduced in \cite{actionness_cvpr14_chen} to indicate the likelihood that a specific spatial location contains a generic action instance. This measure is useful for generating accurate action proposals, as well as for related tasks such as action recognition/detection. It applies to the spatial dimension only though and is thus more like a spatial saliency measure. It works well for spatio-temporal action localization, but is not appropriate for temporal action localization.

Hence, we introduce a new concept, \textit{temporal actionness}, that is better suited to volume data analysis, especially when using 3D ConvNets. Temporal actionness is a fractional value $p_{a}$, ranging from $0$ to $1$, that indicates what proportion of a video volume contains an action of interest temporally. During training, $p_{a}$ is computed as the temporal intersection of a video clip with the actual action period, divided by the clip's length, and is thus a scalar. When $p_{a}$ equals $0$, the volume contains no action. When $p_{a}$ equals $1$, all the frames contain the action. If $p_{a}$ is limited to two values, $0$ or $1$, this problem reduces the standard binary classification problem of action or non-action. 

Compared to the temporal Intersection over Union (tIoU) metric, temporal actionness is more effective because its computation does not depend on the length of ground truth, which means this quantity can be predicted. It based only on how much action of interest a clip contains. On the contrary, predicting the value of tIoU is basically impossible when the ground truth information is missing. 
This may be the primary reason that direct learning of a regressor for time shift, video segment duration, or tIoU usually does not work well in practice.

We design a regression module as shown in Figure \ref{fig:modelConfigure} to learn temporal actionness. 
To demonstrate its effectiveness, we compare it with another approach from \cite{scnn_shou_wang_chang_cvpr16}. That method contains a localization stage with a new loss function and uses tIoU to scale the confidence scores generated by the classification stage. It achieves state-of-the-art performance when training the modules separately. To compare approaches, we simply replace our regression module with their loss function. The results in Section \ref{sec:experiments} show that our proposed model achieves better performance. The benefit of our approach is that an explicit prediction of temporal actionness could also have other applications.


\subsection{Post-processing}
For evaluation, given a video $V$,  we use multi-scale temporal sliding windows to generate input video clips $\{v_{m}\}_{m=1}^{M}$ where $m$ indicates the $m$-th clip and $M$ is the total number of clips for $V$. For each $v_{m}$, we obtain three outputs: the non-action confidence score $p_{b}^{m}$, the action categorization confidence score $\mathbf{p_{l}^{m}}$, and the estimated temporal actionness $p_{a}^{m}$. Bold font indicates vectors.

We design a hierarchical method to integrate these three results. \textit{Clip-level}: For each $v_{m}$, we refine $\mathbf{p_{l}^{m}}$ by multiplying by the temporal actionness value $p_{a}^{m}$. This boosts the categorization confidence scores in clips with lots of action. \textit{Video-level}: For the whole video $V$, we refine $\mathbf{p_{l}^{m}}$ through:
\begin{equation}
	w_{m} \times \mathbf{p_{l}^{m}}, \quad \quad w_{m} = \frac{e^{-\alpha p_{b}^{m}}}{\sum_{n} e^{-\alpha p_{b}^{n}}}
	\label{eq:softmaxWeights}
\end{equation}
where $w_{m}$ is the softmax weights used to scale the categorization confidence scores of the $m$-th video clip \cite{non_action_wang_cvpr16}. In our pipeline, video clips that are predicted as background are not discarded, they are just suppressed. The lower $p_{b}^{m}$, the higher $w_{m}$, which means the prediction scores of a video clip will be increased if it is more likely to be an action clip, and they will be decreased if it is more likely to be a non-action clip. $\alpha$ is a parameter which can be tuned to find the best weighting.

After the clip-level and video-level refinement, we re-score each window using the frequency of its duration in the training set as a per-class duration prior \cite{compact_fisher_action_oneata_iccv13}. However, this re-scoring step may be optional because no substantial improvement was observed. Note that we can further reduce inference time by discarding non-action clips (i.e. when $p_{b}^{m}$ is smaller than some threshold) as most existing works do.
Finally, we apply non-maximum suppression with overlap threshold $\delta$, and select the top-$k$ segments to form the final detection result.

\section{Experiments and Results}
\label{sec:experiments}
In this section, we first describe the implementation details and the benchmark datasets used for evaluation. We then present the experimental results and study the effects of the various design choices described in section \ref{sec:methodology}. 

\subsection{Implementation Details}
For the ConvNets, we use the Keras toolbox.
During training, the network weights are learned using mini-batch ($30$ video clips) stochastic gradient descent with momentum (set to $0.9$). The learning rate is initialized to $0.0001$, except for layer \textsf{fc8} with $0.01$. Both learning rates decrease to $1/10$ of their values whenever the performance saturates, and training is stopped when the learning rate is smaller than $10^{-6}$. Dropout is applied with a ratio of $0.5$. The five losses are fused with equal weights.

For a temporal localization evaluation metric, we follow the standard protocol in THUMOS challenge 2014 to calculate the mean average precision (mAP). A detection is considered to be correct if its overlap with the ground truth action instance is above some threshold, and the action category is predicted correctly.

\subsection{THUMOS Challenge 2014}
THUMOS challenge 2014 is considered to be one of the most challenging datasets for temporal action detection. This dataset consists of $13,320$ trimmed videos from $101$ actions for training, $1,010$ temporally untrimmed videos for validation, and $1,574$ untrimmed videos for testing. The detection task involves $20$ out of the $101$ classes, which are all sports related. During testing, we only use $213$ test videos with annotations to evaluate our framework.

\begin{table}[t]
	\centering
	\scalebox{0.7}{
		\begin{tabular}{ c | c | c | c | c| c }
			\hline
			Model							& 	$\alpha=0.1$		&  $\alpha=0.2$ 	&  $\alpha=0.3$  &  $\alpha=0.4$		&	$\alpha=0.5$ 	\\
			\hline		
			Karaman et al. \cite{florence_thumos2014} 			&   $4.6$ 	& $3.4$	&  $2.1$	&  $1.4$		&  $0.9$			\\
			Wang et al. \cite{motion_appearance_thumos_wang14}		&   $18.2$ 	& $17.0$	&  $14.0$	&  $11.7$		&  $8.3$			\\
			Oneata et al. \cite{learThumos2014}					&   $36.6$ 	& $33.6$	&  $27.0$	&  $20.8$		&  $14.4$			\\
			\hline
			Sun et al. \cite{FGA_web_sun_mm15}  &   $12.4$ 	& $11.0$	&  $8.5$	&  $5.2$		&  $4.4$			\\
			Heilbron et al. \cite{fast_temporal_action_proposal_fabian_cvpr16} &   $36.1$	& $32.9$ & $25.7$	& $18.2$ &  $13.5$		\\
			Richard et al. \cite{action_detection_language_richard_cvpr16}		&   $39.7$ 	& $35.7$	&  $30.0$	&  $23.2$		& $15.2$			\\
			Yeung et al. \cite{frame_glimpse_yeung_cvpr16}	&   $\mathbf{48.9}$ 	& $\mathbf{44.0}$	&  $36.0$	&  $26.4$		&  $17.1$		\\
			\hline
			Ours \textsf{fc8}		&   $45.6$ 	& $41.2$	&  $34.5$	&  $26.1$		&  $17.0$		\\	
			Ours \textsf{conv5} + \textsf{fc8}		&   $46.6$ 	& $42.9$	&  $35.6$	&  $28.5$		&  $18.7$		\\	
			Ours \textsf{conv5} + \textsf{fc6} + \textsf{fc8}	&   $47.7$ 	& $43.6$	&  $\mathbf{36.2}$	&  $\mathbf{28.9}$		&  $\mathbf{19.0}$	\\	
			\hline
		\end{tabular}
	}
	\caption{Action detection results on THUMOS 2014 with varied IoU threshold $\alpha$. All performances are reported using mAP. Top section: Top three performers in the 2014 challenge. Middle section: Recent state-of-the-art approaches.  Bottom section: Our approach. We observe that our network with multiple intermediate losses performs the best. Note that \textsf{conv5}, \textsf{fc6}, \textsf{fc8} here refer to losses not features. \label{tab:thumos14Result}}
	\vspace{-2ex}
\end{table} 

\noindent \textbf{Results:}
Since our goal is temporal action localization for untrimmed videos, we include the validation videos in our training data. We have already described how we generate temporal multi-scale video clips for untrimmed videos in Section \ref{sec:methodology}. For trimmed videos, we uniformly sample $16$ frames as clips without multi-scale augmentation, and each clip has the same action label as the video. 
During training, we perform $15$K iterations with learning rate $10^{-4}$, $15$K iterations with $10^{-5}$, and $10$K iterations with $10^{-6}$. During prediction, the overlap threshold in NMS is set to $0.4$.

Table \ref{tab:thumos14Result} shows that our method outperforms previous state-of-the-art approaches.
The three models in the top section of Table \ref{tab:thumos14Result} use Fisher vector encoded IDT features on video clips and perform post-processing to refine the localization scores. \cite{learThumos2014} is the winning approach of the 2014 challenge. It uses video-level action classification scores as a context feature which greatly improves the performance.
The four methods in the middle section of Table \ref{tab:thumos14Result} are the most recent state-of-the-art results. 
Note that \cite{frame_glimpse_yeung_cvpr16} achieves better accuracy in the small $\alpha$ regime. One possible reason is that their learned observation policy tends to output shorter segments compared to a standard sliding window approach.
Significantly, though, we estimate that the computation time required to train the approach of \cite{frame_glimpse_yeung_cvpr16} is much higher than ours. Their observation network, a fine-tuned VGG-16 model, by itself requires approximately the same amount of computation time to train as our single 3D ConvNets. They also have to train the RNN-based agent to learn the decision policy.
\cite{scnn_shou_wang_chang_cvpr16} reports a mAP of $19.0$ when $\alpha = 0.5$. This is using the updated evaluation toolkit \cite{THUMOS14}, however, and so is not directly comparably with  the results in Table \ref{tab:thumos14Result}. For completeness, we also evaluate our result using the updated toolkit and obtain an accuracy of $19.2$ given our multi-loss fusion network. 
Furthermore, our approach is several times faster to train/evaluate than \cite{scnn_shou_wang_chang_cvpr16} and we only need to store one model instead of three. More importantly, our method is simpler to train than the multi-stage pipeline of \cite{scnn_shou_wang_chang_cvpr16} which requires manually balanced datasets and separate input files for the three stages.

\begin{figure}[t]
	\centering
	\includegraphics[width=1.0\linewidth,trim=0 0 0 0,clip]{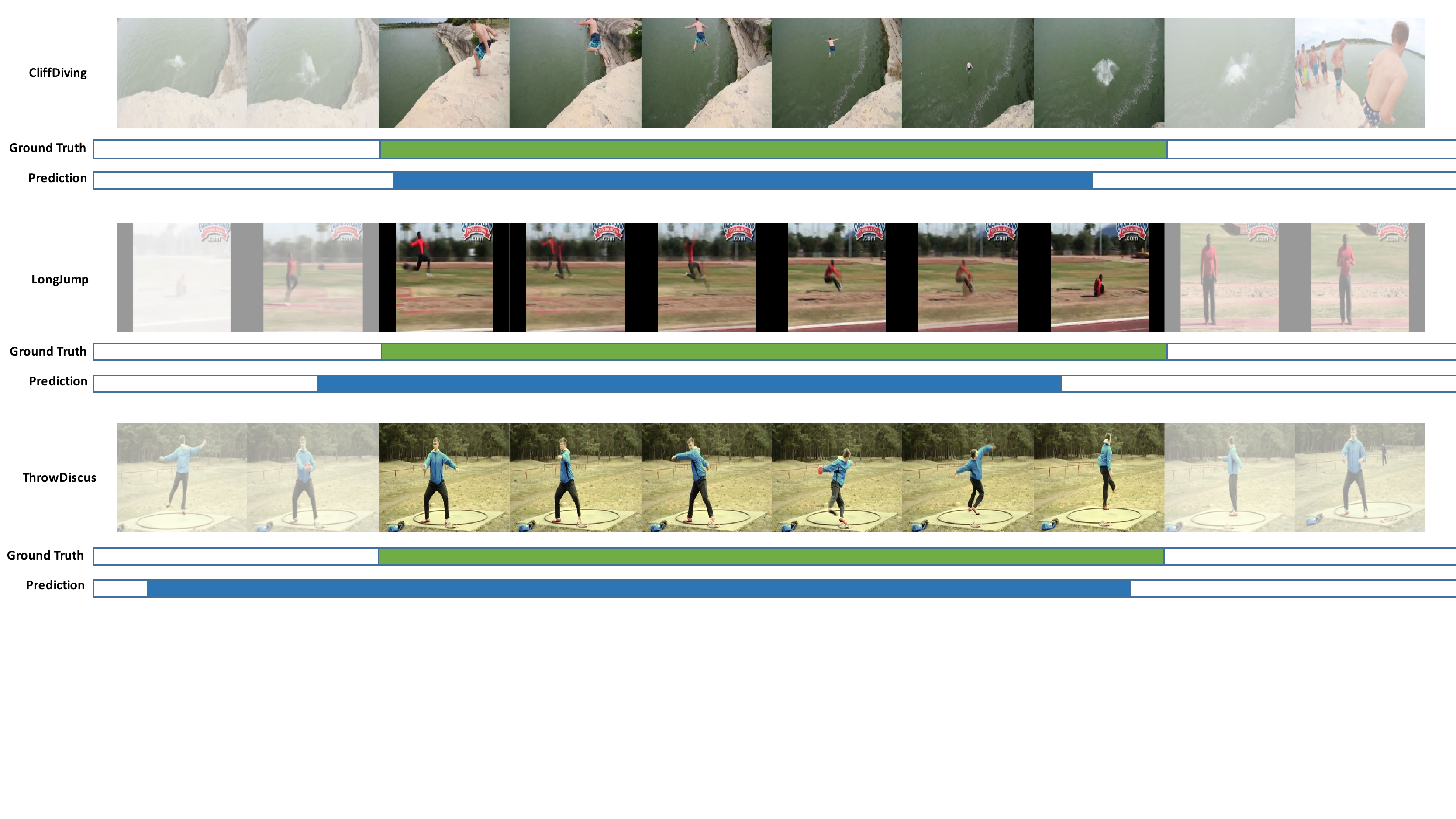}
	\vspace{-10ex}
	\caption{Examples of predicted action instances on THUMOS 2014. Faded frames indicate background. Ground truth is green while our predictions are blue. Top: CliffDiving. Middle: LongJump. Bottom: ThrowDiscus. This figure is best viewed in color. }
	\label{fig:thumosdemo}
\end{figure}

Figure \ref{fig:thumosdemo} shows three examples of predicted action instances from the THUMOS 2014 challenge test set. Our approach is shown to accurately localize the action instances. 

\begin{table}[t]
	\begin{minipage}{0.25\textwidth}%
		\centering
		\subfloat[][Sequential Network \cite{scnn_shou_wang_chang_cvpr16}]{
			\scalebox{0.7}{
				\begin{tabular}{| c | c | c | }
					\hline
					Training Set							&  $\alpha=0.2$ 	&	$\alpha=0.5$ 	\\
					\hline		
					$V_{T}$ + $V_{U}$ 						  	& $43.5$	&  $19.0$		\\	
					$\frac{3}{4} V_{T}$ + $V_{U}$ 			  & $41.7$	  	  &  $17.9$			\\
					$\frac{1}{2} V_{T}$ + $V_{U}$  			  & $36.9$	 &  $14.4$		\\	
					\hline
				\end{tabular}
			}
		}
	\end{minipage}%
	\begin{minipage}{0.25\textwidth}%
		\centering
		\subfloat[][Our Parallel Network]{
			\scalebox{0.7}{
				\begin{tabular}{| c | c | c | }
					\hline
					Training Set							&  $\alpha=0.2$ 	&	$\alpha=0.5$ 	\\
					\hline		
					$V_{T}$ + $V_{U}$ 						  	& $43.6$	&  $19.2$		\\	
					$\frac{3}{4} V_{T}$ + $V_{U}$ 			  & $42.9$	  	  &  $18.7$			\\
					$\frac{1}{2} V_{T}$ + $V_{U}$  			  & $39.4$	 &  $17.3$		\\	
					\hline
				\end{tabular}
			}
		}
	\end{minipage}%
	\caption{Comparison between our parallel multi-task learning pipeline and its sequential counterpart in \cite{scnn_shou_wang_chang_cvpr16}. Our method is robust to large reductions in training data. Note the mAP here is reported using the updated evaluation toolkit.\label{tab:regularization}}
\end{table}


\subsection{Discussion}
In this subsection, we will discuss the benefits of using our parallel learning framework over its sequential counterpart. We choose THUMOS 2014 as the illustrating dataset. 

\noindent \textbf{Enhanced Regularization}
Multi-task learning can benefit from the enhanced regularization provided by closely-related tasks. This can increase a model's robustness, especially under the fewer training samples regime. Here, we explore this advantage of our joint learning scheme by gradually reducing the number of trimmed videos in the training set. Our training set for THUMOS 2014 is composed of two parts, the original trimmed training videos which we denote as $V_{T}$, and the untrimmed validation videos which we denote as $V_{U}$. By decreasing the proportion of $V_{T}$ in our training set, we can validate our framework's robustness.

As shown in Table \ref{tab:regularization}(b), we achieve the best performance using the full training dataset, as expected. Notably, though, the performance decreases only slightly when just $3/4$ of $V_{T}$ is used. The relative reduction is still only $9.8\%$ when $1/2$ of $V_{T}$ is used. 

By contrast, less training data is much more likely to result in overfitting in a multi-stage approach like \cite{scnn_shou_wang_chang_cvpr16}. To demonstrate this, we trained the second ``action categorization'' stage of \cite{scnn_shou_wang_chang_cvpr16} using $1/2$ of $V_{T}$ and all $V_{U}$ and the classification accuracy of the 3D ConvNets dropped dramatically. This, in turn, will degrade the performance of the whole system since the localization network is initialized from the categorization stage. The final action detection mAP is $14.4$ as shown in Table \ref{tab:regularization}(a). The relative performance decrease is $24.2\%$ for the sequential framework. This demonstrates that our joint learning framework is more robust when fewer trimmed video examples are available. This is important for video analysis because current video datasets do not have large numbers of labeled training samples. 

\noindent \textbf{Computational and Storage Efficiency}
Our approach is computationally efficient compared to approaches that compute and fuse expensive features \cite{learThumos2014} or to approaches with multiple stages \cite{scnn_shou_wang_chang_cvpr16,fast_temporal_action_proposal_fabian_cvpr16,frame_glimpse_yeung_cvpr16}. Our single model only requires $40K$ iterations to train and each iteration takes approximately one second. Although non-action video clips are not discarded, our class-balanced sampling strategy effectively controls how many clips our ConvNet sees during training. For THUMOS 2014, our model sees less than $10\%$ of the video frames. 

For test time efficiency, our network is almost two times faster than the multi-stage pipeline \cite{scnn_shou_wang_chang_cvpr16}. Intuitively, although \cite{scnn_shou_wang_chang_cvpr16}'s proposal stage can discard many video segments before the classification stage, all the slided windows need to go through a 3D ConvNet first, and then the remaining candidates go through a classification 3D ConNet one more time. Our multi-tasking network only require all the slided windows to go though a 3D ConvNet once. 
We can further increase test time efficiency by controlling the proportion of non-action video clips to preserve by setting a threshold. 

With regards to storage, our approach does not need to store high-dimensional representations, such as Fisher Vector encoded features. And, only a single ConvNet model is required for deployment whose size is less than $0.35$GB. 

\begin{table}[t]
	\begin{minipage}{0.24\textwidth}%
		\centering
		\subfloat[][]{
			\scalebox{0.7}{
				\begin{tabular}{| c | c | c | }
					\hline
					Data Balance						&  $0.2$ 	&	$0.5$ 	\\
					\hline		
					Proposal 						  	& $40.1$	&  $17.9$		\\	
					Categorization			  		& $43.6$	  	  &  $19.0$			\\
					\hline
				\end{tabular}
			}
		}
	\end{minipage}%
	\begin{minipage}{0.24\textwidth}%
		\centering
		\subfloat[][]{
			\scalebox{0.7}{
				\begin{tabular}{| c | c | c |}
					\hline
					Overlap							&  $0.2$ 	&	$0.5$ 	\\
					\hline		
					Localization loss \cite{scnn_shou_wang_chang_cvpr16} 						  	& $43.2$	&  $18.9$		\\	
					Temporal actionness & $43.6$	 &  $19.0$		\\	
					\hline
				\end{tabular}
			}
		}
	\end{minipage}%
	\vspace{1ex}
	\begin{minipage}{0.24\textwidth}%
		\centering
		\subfloat[][]{
			\scalebox{0.7}{
				\begin{tabular}{ | c | c | c |}
					\hline
					Shear Param			&  $0.2$ 	&	$0.5$ 	\\
					\hline		
					$\theta = 0$ 			  & $40.6$	&  $18.3$		\\	
					$\theta = 10$ 			  & $40.9$	&  $18.3$		\\	
					$\theta = 25$ 			  & $43.6$	  	  &  $19.0$			\\
					$\theta = 40$  			  & $39.7$	 &  $17.8$		\\	
					\hline
				\end{tabular}
			}
		}
	\end{minipage}%
	\begin{minipage}{0.24\textwidth}%
		\centering
		\subfloat[][]{
			\scalebox{0.7}{
				\begin{tabular}{ | c | c | c |}
					\hline
					Contribution							&  $0.2$ 	&	$0.5$ 	\\
					\hline		
					w/o proposal						  	& $41.7$	&  $18.2$		\\	
					w/o regression 			  & $40.0$	 &  $17.5$		\\	
					full model						  	& $43.6$	&  $19.0$		\\	
					\hline
				\end{tabular}
			}
		}
	\end{minipage}%
	\caption{Action detection results of our framework on THUMOS 2014 under different configurations. (a) Data imbalance during training. (b) Effectiveness of proposed temporal actionness. (c) Random shear augmentation. (d) Impact of each branch. See the text in Ablation Study for more details. \label{tab:Ablation}}%
	\vspace{-2ex}
\end{table} 

\noindent \textbf{Ablation Study} 
We conduct ablative experiments to understand the effects of the different components and parameters in our framework.

First, we compare different strategies for balancing the training data. As shown in Table \ref{tab:Ablation}(a), balancing for the action categorization task yields better performance, especially when $\alpha=0.2$. As mentioned before, balancing for the action proposal task likely results in the network seeing too many negative examples and features that are tuned more to background video clips than to actions of interest. 

In Table \ref{tab:Ablation}(b), we replace our regression module based on temporal actionness with the well-designed localization loss function proposed in \cite{scnn_shou_wang_chang_cvpr16}. Our approach obtains slightly better performance, but note the temporal actionness explicitly predicted by our network can have other applications.

We also determine the optimal value of the shear parameter $\theta$ from among $10$, $25$ and $40$ degrees. In Table \ref{tab:Ablation}(c), $\theta=0$ means there is no shear augmentation. When $\theta=10$, the performance does not change. One possible reason is that the corner-cropping augmentation might already provide similar benefits. When $\theta=40$, the performance decreases by $0.5\%$, possibly due to too much distortion. For $\theta=25$, we observe an improvement of $0.7\%$ when $\alpha=0.5$. Considering that the best mAP reported is $19.0$ when tIoU is $0.5$, an improvement from random shear of $0.7\%$ is relatively significant. Most importantly, random shear improves the accuracy of the ``classification'' branch by $1.6\%$ on THUMOS 2014. After the post-processing and NMS, the improvement reduces to $0.7\%$ for the final action detection. Hence, random shear augmentation is very useful to help overcome overfitting during model training. 


Finally, we explore whether our action proposal and temporal actionness branches are redundant since the former can be seen as an extreme case of the latter. As shown in Table \ref{tab:Ablation}(d), the full model achieves the best performance, indicating the two branches are complementary. The \textit{w/o proposal} model performs $0.8\%$ worse than the full model, while the \textit{w/o regression} model performs $1.5\%$ worse. This demonstrates that estimating temporal actionness is more beneficial than performing action proposal.

\begin{table}[t]
	\begin{minipage}{0.25\textwidth}%
		\centering
		\subfloat[][MSR Action II]{
			\scalebox{0.7}{
				\begin{tabular}{ | c | c | }
					\hline
					Model								&	$\alpha=0.5$ 	\\
					\hline
					Yu et al. \cite{yu_FAP_cvpr15} &  $28.2$	\\
					Gemert et al. \cite{APT_Gemert_bmvc15} &  $54.5$\\
					Heilbron et al. \cite{fast_temporal_action_proposal_fabian_cvpr16} 	&  $60.3$		\\
					\hline
					Ours + \textsf{fc8}		&  $59.6$		\\	
					Ours \textsf{conv5} + \textsf{fc6} + \textsf{fc8}  &  $\mathbf{61.1}$		\\	
					\hline
				\end{tabular}
			}
		}
	\end{minipage}%
	\begin{minipage}{0.25\textwidth}%
		\centering
		\subfloat[][MPII Cooking]{
			\scalebox{0.7}{
				\begin{tabular}{| c | c | }
					\hline
					Model								&	$\alpha=0.5$ 	\\
					\hline
					Sliding Window  &  $7.9$	\\
					Gemert et al. \cite{APT_Gemert_bmvc15} &  $13.1$\\
					Richard et al. \cite{action_detection_language_richard_cvpr16} 	&  $14.0$		\\
					\hline
					Ours + \textsf{fc8}		&  $14.6$		\\	
					Ours \textsf{conv5} + \textsf{fc6} + \textsf{fc8}		&  $\mathbf{14.9}$		\\	
					\hline
				\end{tabular}
			}
		}
	\end{minipage}%
	\caption{Action detection results on MSR Action Dataset II and MPII Cooking dataset. Our approach consistently achieves better performance compared to state-of-the-art approaches at $0.5$ overlap threshold.\label{tab:MSRAndMEX}}%
\end{table} 

\subsection{MSR Action Dataset II}
The MSR Action Dataset II \cite{MSR_action_II_PAMI11} consists of $54$ video sequences ($203$ action instances) recorded in a crowded environment. There are three action types: hand waving, handclapping, and boxing. We follow the standard cross-dataset evaluation protocol of using the KTH dataset for training.  During training, we perform $8$K iterations with learning rate $10^{-4}$, $8$K iterations with $10^{-5}$, and $4$K iterations with $10^{-6}$. During prediction, the overlap threshold in NMS is $0.5$.

Table \ref{tab:MSRAndMEX}(a) shows that our joint learning framework with multiple intermediate losses performs better than the state-of-the-art approach of \cite{fast_temporal_action_proposal_fabian_cvpr16}. That method uses IDT features to represent the video clips, however, which prevents it from being applied to large-scale datasets since the Fisher vector encoded features are too large to be cached. Our approach can easily scale to large datasets. We attribute our good performance, at least in part, to the joint regularized learning and to the temporal actionness refined post-processing. Note that, \cite{yu_FAP_cvpr15} and \cite{APT_Gemert_bmvc15} are designed to generate spatio-temporal proposals. \cite{fast_temporal_action_proposal_fabian_cvpr16} projects these proposals to the temporal dimension and randomly picks a subset of temporal proposals to perform action detection. An example of predicted ``boxing'' instance using our approach can be seen in Figure \ref{fig:MSRandMPII} top.

\subsection{MPII Cooking}
MPII Cooking \cite{MPII_cooking_cvpr12} is a large, fine-grained cooking activities dataset. It contains $44$ videos with a total length of more than $8$ hours of $12$ participants performing $65$ different cooking activities. It consists of a total of 5,609 annotations spread over the $65$ activity categories, including a background class for the action detection task. It also has a second type of annotation, articulated human pose for a pose estimation problem, which we ignore. Following the standard protocol in \cite{MPII_cooking_cvpr12}, we have $7$ splits after performing leave-one-person-out cross-validation. Each split uses $11$ subjects for training, leaving one for validation. During training, we perform $10$K iterations with learning rate $10^{-4}$, $10$K iterations with $10^{-5}$, and $10$K iterations with $10^{-6}$. During prediction, the overlap threshold in NMS is set to $0.5$.

\begin{figure}[t]
	\centering
	\includegraphics[width=1.0\linewidth,trim=0 0 0 0,clip]{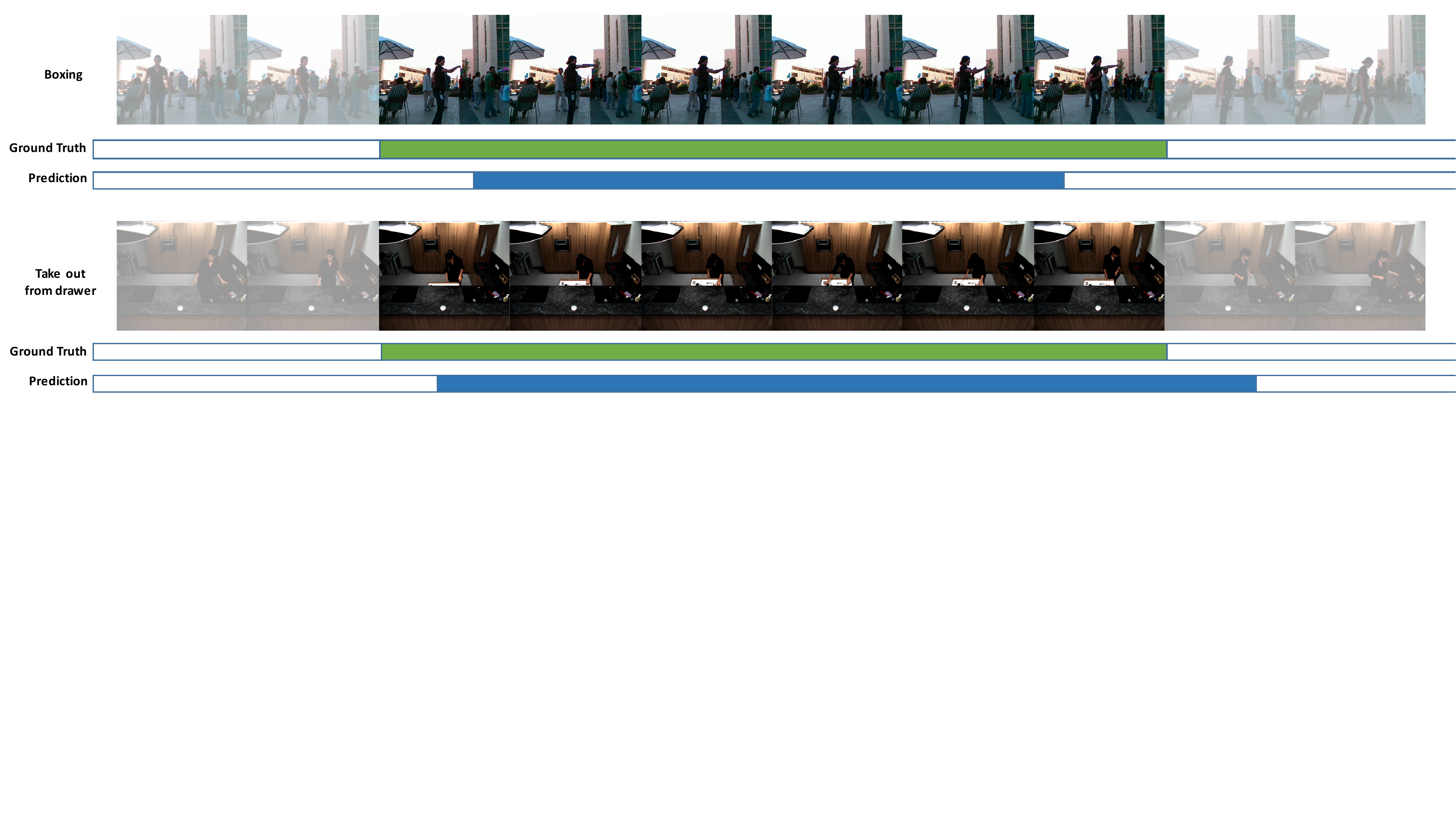}
	\vspace{-18ex}
	\caption{Examples of predicted action instances. Faded frames indicate background. Ground truth is green while our predictions are blue. Top: ``Boxing'' in MSR Action Dataset II. Bottom: ``Take out from drawer'' in MPII Cooking dataset. This figure is best viewed in color. }
	\label{fig:MSRandMPII}
\end{figure}

We compare our method to a sliding window baseline similar to \cite{action_detection_language_richard_cvpr16,MPII_cooking_cvpr12}.
Table \ref{tab:MSRAndMEX}(b) shows our method performs better than both the baseline and recent state-of-the-art approaches \cite{action_detection_language_richard_cvpr16,APT_Gemert_bmvc15}. \cite{action_detection_language_richard_cvpr16} is novel in that it includes a length and language model in addition to an action classifier. 
However, their runtime during inference is quadratic in the number of frames. By limiting the maximal action length to a constant, they can solve the action detection problem in a reasonable time, but this is not easily scalable to long videos. 
When we compare our approach to \cite{APT_Gemert_bmvc15}, we report their performance by applying the implementation kindly provided by the authors. An example of predicted ``Take out from drawer'' instance using our approach can be seen in Figure \ref{fig:MSRandMPII} bottom.

\section{Conclusion}
\label{sec:conclusion}
We present a novel multi-task learning framework to detect human actions in long, untrimmed videos. 
Performing the three related tasks in parallel exploits the training data more thoroughly and is more efficient in time and space. The random shear augmentation simulates viewpoint differences. The multiple intermediate losses force the network to learn better representations for action categorization. These theoretical insights are supported by our experimental results on three popular datasets.
In future work, we plan to revise our multi-task approach to use a 3D residual learning framework \cite{residual_cvpr16}. 
We are also interested in optimizing the network end-to-end similarly to Faster R-CNN \cite{fasterRCNN}, since our temporal actionness scores can be considered as region proposal scores. 
\newline

\noindent \textbf{Acknowledgements.} This work was funded in part by a National Science Foundation CAREER grant, $\#$IIS-1150115, and a seed grant from the Center for Information Technology in the Interest of Society (CITRIS). We gratefully acknowledge NVIDIA Corporation through the donation of the Titan X GPU used in this work.

{\small
	\bibliographystyle{ieee}
	\bibliography{egbib_cvpr}
}

\end{document}